\documentclass[letterpaper]{article}
\usepackage{aaai19}
\usepackage{times}
\usepackage{helvet}
\usepackage{courier}
\usepackage{url}
\usepackage{graphicx}
\usepackage{algorithm}
\usepackage[noend]{algpseudocode}
\usepackage{times}
\usepackage{comment}
\usepackage{amsmath}
\usepackage{amsfonts}

\newtheorem{definition}{Definition}[section]
\newcommand{\RNum}[1]{\uppercase\expandafter{\romannumeral #1\relax}}

\title{Learning Vine Copula Models For Synthetic Data Generation}
\author{
Yi Sun  \\
MIT \\
Cambridge, MA
\And 
Alfredo Cuesta-Infante\\
Universidad Rey Juan Carlos \\
Madrid, Spain
\And Kalyan Veeramachaneni \\
MIT \\
Cambridge, MA
}
\begin{document}
\maketitle
\begin{abstract}
  A vine copula model is a flexible high-dimensional dependence model which uses only bivariate building blocks. However, the number of possible configurations of a vine copula grows exponentially as the number of variables increases, making model selection a major challenge in development. In this work, we formulate a vine structure learning problem with both vector and reinforcement learning representation. We use neural network to find the embeddings for the best possible vine model and generate a structure. Throughout experiments on synthetic and real-world datasets, we show that our proposed approach fits the data better in terms of log-likelihood. Moreover, we demonstrate that the model is able to generate high-quality samples in a variety of applications, making it a good candidate for synthetic data generation.  
\end{abstract}

\section{Introduction}
\label{sec:Intro} 

The machine learning (ML) community is increasingly interested generative modeling. Broadly, generative modeling consists of modeling either both the joint distribution of data and classes for supervised learning or of modeling only the joint distribution of data for unsupervised learning. In tasks involving classification, a generative model is useful to augment smaller, labeled datasets, which are especially problematic when developing deep learning applications for novel fields \cite{jason_2017}.

In unsupervised learning (clustering), a generative model supports development of model-driven algorithms (where each cluster is represented by a model). These algorithms scale better than data-driven algorithms, and are typically faster and more reliable. Additionally, these algorithms are crucial tools in tasks that don't easily fit in the supervised/unsupervised paradigm, including, for example, survival analysis - e.g. predicting when a chronically ill person will return to the hospital, or how long will a project last in a Kickstarter \cite{Vinzamuri_2014}.

Synthetic data generation - i.e. sampling new instances from joint distribution - can also be carried out by a generative model. Synthetic data has found multiple uses within machine learning. On one hand, it is useful for testing the scalability and the robustness of new algorithms; on the other, it is safe to be shared openly, preserving the privacy and confidentiality of the actual data \cite{li2014dpsynthesizer}. 

The central underlying problem for generative modeling is to construct a joint probability distribution function, usually high-dimensional and comprising both continuous and discrete random variables. This is often accomplished by using probabilistic graphical models (PGM) such as Bayesian networks (BN) and conditional random fields (CRF), to mention only two of many possible approaches \cite{Jordan_1999,Koller_Friedman_2009}. In PGM, the joint probability distribution obtained is simplified due to assumptions on the dependence between variables, which is represented in form of a graph. Thus the major task for PGM is the process of learning such a graph; this problem is often understood as a structure learning task that can be solved in a constructive way, adding one node at a time while attempting to maximize the likelihood or some information criterion. In PGM, continuous variables are quantized before a structure is learned or parameters identified, and, due to this quantization, this solution looses information and scales poorly. 

Copula functions are joint probability distributions in which any univariate continuous probability distribution can be plugged in as a marginal. Thus, the copula captures the joint behaviour of the variables and models the dependence structure, whereas each marginal models the individual behaviour of its corresponding variable. In other words, the choice of the copula and the marginals results in the construction of the joint probability  distribution directly \cite{nelsen_2006}. However, in practice, there are many bivariate copula families but only a few multivariate ones, with the Gaussian copula and the T-copula being the most prominent. For this reason, these two families have been used extensively, leading to models that most of the time outperform the multivariate normal (MVN). However, these models still assume a dependence structure that may only loosely capture the interaction between a subset of variables. The strongest and most well-known case against the abuse of Gaussian copula can be observed in \cite{MacKenzie_Spears_2014}. The key problem pinpointed in that work is that financial quantities are seldom jointly linear, and even if so, the measure of such association, i.e. the correlation, is not stable across time.
Therefore, the Gaussian bivariate copula, whose parameter is the correlation between its covariates, is a bad choice.
Other copula families depend on non-linear degrees of association such as Kendall's $\tau$, but it is equally unwise to model the joint behavior of more than two covariates with any of them on the hope that a single scalar will be able to capture all the pairwise dependencies.

Vine copulas provide a powerful alternative in modeling dependence of high-dimensional distributions \cite{kurowicka_joe_2011}. To explicate, consider the same case presented in \cite{MacKenzie_Spears_2014} under three distinct generative models. Let the multivariate normal distribution (MVN) be the first one, the Gaussian copula next, and finally a vine model. For the sake of simplicity, let us also assume that there are three covariates involved, $x_i$, for $i=\{1,2,3\}$ with probability density functions (PDF) $f_i(x_i)$. Hence, the MVN correlation matrix $\mathbf{R} \in [-1,1]^{3\times 3}$. Since the marginals of MVN are also normal whereas the actual marginals $f_i$ might be very different, samples from the MVN are very likely to represent the actual data poorly. In this case, one would then proceed by finding $\hat{f}_i(x_i)$, the best match with the actual marginals ${f}_i(x_i)$. Together with the Gaussian copula, a more accurate model is thus constructed and sampled. If covariates are not jointly linear, samples will differ from actual data;
for example, when $x_1$ and $x_2$ occur jointly in similar rank positions, but not necessarily in linear correlation, Kendall's $\tau$ rank correlation is a much better parameter, and this pair of covariates would be better modeled by a copula parameterized by $\tau$, such as the Clayton, the Frank or the Gumbel family, to mention only the most popular ones. This copula would be the first block of the vine model. 
Following with the example, next we have to plug the third covariate to either the left end or the right end of the tree, the decision is due to some metric such as likelihood, information criteria, goodness-of-fit, etc,  with another copula following the same procedure. Thus, the first tree for three covariate would be ended. 
For the second tree, the copulas (edges) in the previous trees are now nodes that are to be linked. Since we only have two, they can only form one possible bond, and the construction ends for we cannot iterate once more.


The dependence structure in Vine copulas is constructed using bivariate copulas as building blocks; thus, two or more variables can have a completely different behavior than the rest. A vine is represented as a sequence of trees, organized in levels, so they can be considered as a PGM. Another distinctive feature of vine copulas is that the joint probability density factorization resulting from the entire graph can be derived by the chain rule. In other words, theoretically, there are no assumptions about the independence between pairs of variables, and in fact the building block in such a case is the independence copula. However, in practice usually only the top levels of a vine are constructed. Therefore, learning a vine has the cost of learning the graph, tree-by-tree, plus the cost of selecting the function that bonds each pair of nodes. 

The problem is more challenging when the vine is pruned from one level downwards to the last. Yet this effort is rewarded by a superior and more realistic model. 


This paper presents the following contributions. Firstly, we formulate the model selection problem for regular vine as a reinforcement learning (RL) problem and relax the tree-by-tree modeling assumption, where each level of tree is selected sequentially. Moreover, we use long-short term  memory (LSTM) networks in order to learn from vine configurations tested long and short ago. Second, as far as we are aware, this work is the first to use regular vine copula models for generative modeling and synthetic data generation. Finally, a novel and functional technique for evaluating model accuracy is a side result of synthetic data generation. We propose that a model can be admitted if it generates data that produces a performance similar to the actual data on a number of ML techniques; e.g. decision trees, SVM, Neural Networks, etc.

The rest of the paper is organized as follows. In section ~\ref{sec:Related}, we present the literature related to the topics of this paper. In section ~\ref{rvine}, we introduce the definition and construction of a regular vine copula model. Section ~\ref{method} describes our proposed learning approach for constructing regular vine model. In section ~\ref{experiments}, we apply the algorithm to several synthetic and real datasets and evaluate the model in terms of fitness and the quality of the samples generated.

\section{Motivation and Related work}
\label{sec:Related}

The rise of deep learning (DL) in the current decade has brought forth new machine learning techniques such as convolutional neural networks (CNN), long-short term memory networks (LSTM) or generative adversarial networks (GAN) \cite{Goodfellow_etal_2016}. These techniques outrank the state-of-the-art in problems from many fields, but require large datasets for training, which can be a significant problem given that often collecting data is expensive or time consuming. Even when data is already collected, often it cannot be released due to privacy or confidentiality issues. Synthetic data generation, currently a well researched topic in machine learning, provides a promising solutions for these problems \cite{Alzantot_etal_2017,Libes_2017,Soltana_etal_2017,Sagduyu_etal_2018}.
Generative models - that is, high-dimensional multivariate probability distribution functions - are a natural way to generate data. More recently, the rise of GAN and its variations provides a way of generating realistic synthetic images \cite{Goodfellow_etal_2016,alex_2017}.

Copula functions, and PGM involving copulas and vines, have gained momentum in ML since the early proposal of the copula based regression models \cite{Kolev_Paiva_2009} and copula Bayesian networks \cite{Elidan_2010}. Gaussian process vine copulas were introduced in \cite{LopezPaz_etal_2013a}. The Copula Discriminant Analysis (CODA) is a high-dimensional classification method based on the Gaussian Copula proposed in \cite{Han_etal_2013}. The multi-task copula was introduced in \cite{Zhou_Tao_2014} for multi-task learning. A copula approach has been applied to jointly modeling of longitudinal measurements and survival times in AIDS studies \cite{Ganjali_2015}. A vine copula classifier has performed competitively compared to the four best classification methods presented at the Mind Reading Challenge Competition 2011 \cite{Carrera2016}. 

In this paper we obtain them by means of a copula function based PGM known as Vine. 
Vines were first introduced in \cite{Bedford_Cooke_2001} as a probabilistic construction of multivariate distributions based on the simple building blocks of bi-variate copulas. These constructions are organized  graphically as a sequence of nested undirected trees. Compared to black-box deep learning models, vine copula has better interpretability since it uses a graph like structure to represent correlations between variables. However, learning a vine model is generally a hard problem.
In general, there exists $\frac{d!}{2}2^{\binom{d-2}{2}}$ different $d$-dimensional regular vines with $d$ variables, and $|B|^{\binom{d}{2}}$ different combinations of bivariate copula families where $|B|$ is the size of the candidate bivariate families \cite{MoralesNapoles_2010}. 

To reduce the complexity of model selection, \cite{DiBmann_2013} proposed a tree-by-tree approach that selects each tree $T_1,...,T_{d-1}$ sequentially, with a greedy maximum-spanning tree algorithm where edge weights are chosen to reflect large dependencies. 
Although this method works well in lower-dimensional problems, the greedy approach does not ensure optimal solutions for high-dimensional data.
Gruber and Czado (\citeyear{Gruber_Czado_2015}, \citeyear{Gruber_Czado_2018})  proposed a Bayesian approach to estimate regular vine structure along with the pair copula families from an arbitrary set of candidate families. 
However, the sequential, tree-by-tree, Bayesian approach is computationally intensive and cannot be used for more than 20 dimensions.
A novel approach to high-dimensional copulas has been recently proposed by M\"uller and Czado (\citeyear{Muller_Czado_2018}). 

In this paper, we reformulate the model selection as a sequential decision-making process, and cast it as an RL problem, which we solve  with policy learning \cite{Sutton_Barto_98}. 
Additionaly, when constructing the Vine, a decision made in the first tree can limit the choices in construction of subsequent trees. Therefore, we cannot assure the markovian property, i.e. that the next state depends only of the current state and current decisions. Such a non-markovian property suggests the use of Long Short Term Memory (LSTM) \cite{sepp_1997}. LSTM in conjunction with model-free RL was presented in \cite{Bakker_2001} as a solution to non-Markovian RL tasks with long-term dependencies. 

\section{Regular Vine Copula}\label{rvine} %

\begin{figure}[t]
    \centering
    \includegraphics[scale=0.25]{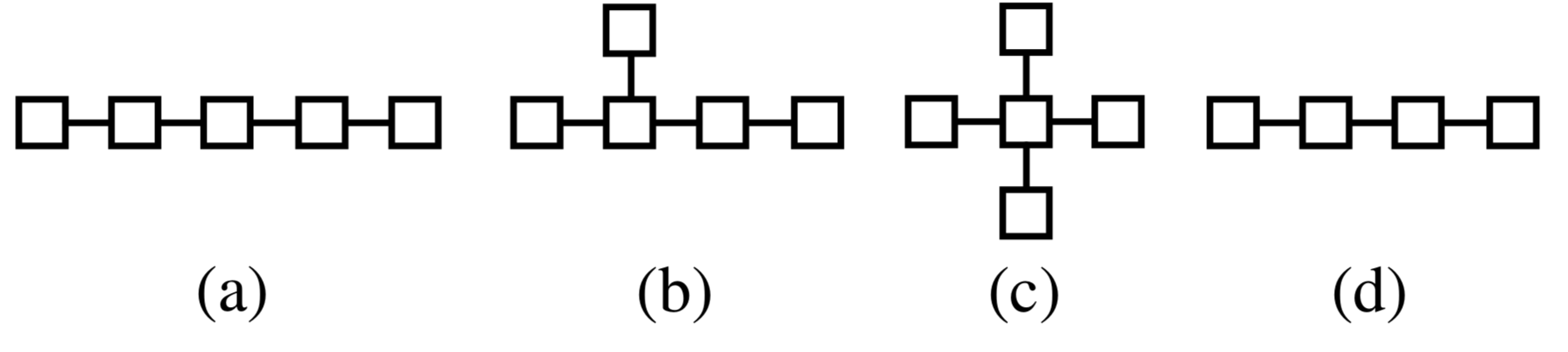}\\
    \includegraphics[scale=0.25]{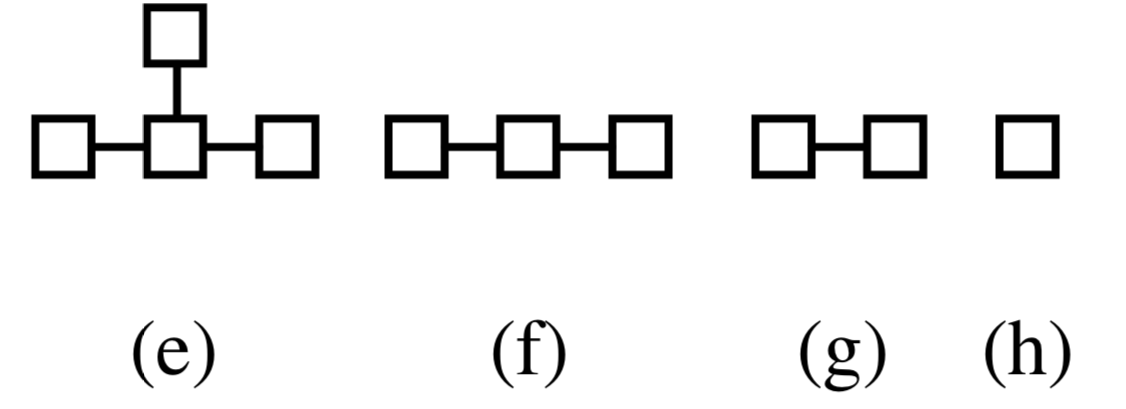}
    \caption{(a-h) All the different layouts of every possible tree in a 5-dim vine. (a-c) correspond to the 1st level, (d-e) correspond to the 2nd level, (f,g,h) are the unique possible layouts for trees in 3rd, 4th and 5th level respectively. }
    \label{fig:vine_layouts}
\end{figure}

\begin{figure}[t]
\centering
    \includegraphics[scale=0.23]{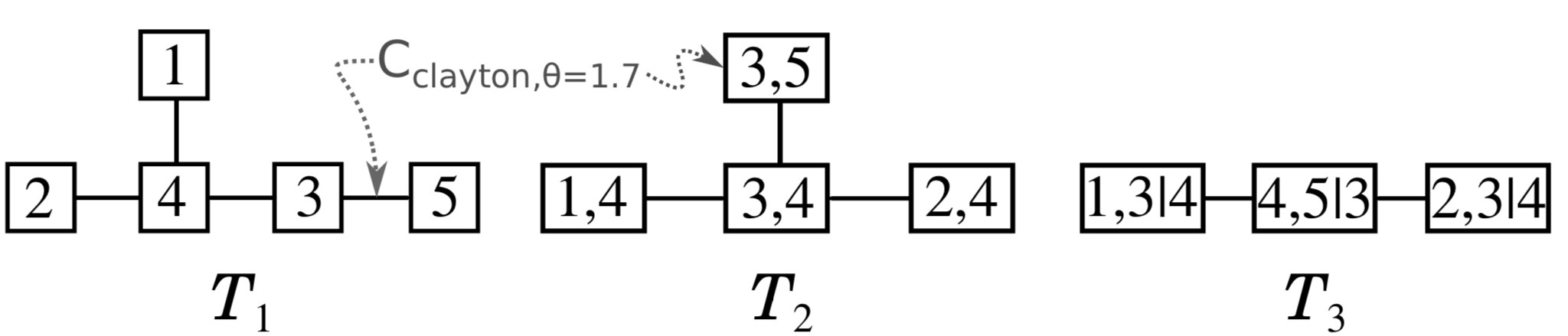}\\
    \includegraphics[scale=0.23]{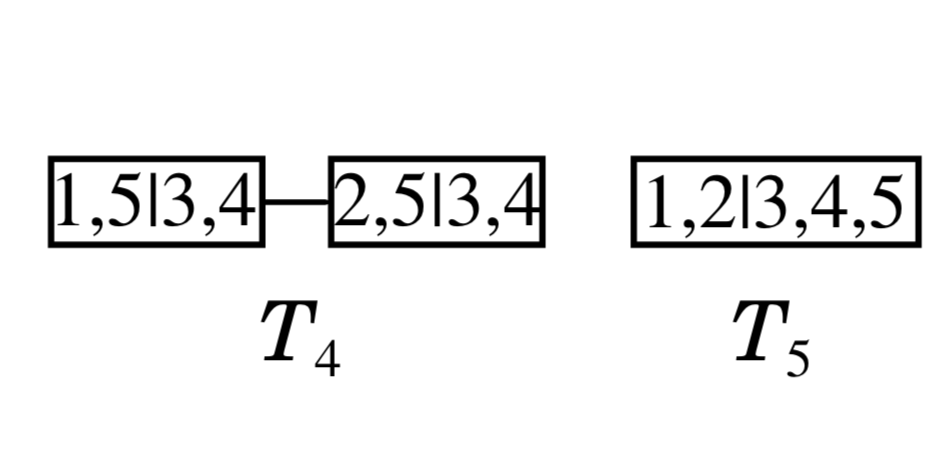}
    \caption{An example of 5-dim vine constructed with layouts \{b,e,f,g,h\} in Figure \ref{fig:vine_example}, and a valid choice of edges in it. For the sake of clarity, there is only one edge with the detail of its copula family and parameter.
    }
    \label{fig:vine_example}
\end{figure}

In this section we summarize the essential facts about the meaning of the vine graphical model and how it is transformed into a factorization of the copula density function that models the dependence structure.
A deeper explanation about vines can be found in \cite{Aas_etal_2009}.

According to Sklar's theorem, the copula density function is what it needs to complete the joint probability density of $d$ continuous covariates in case they are not independent. 
In other words, if the individual behaviour of covariate $x_i$ is given by the marginal probability density function $f_i(x_i)$ for $i=1,\ldots,d$, then the copula $c$ brings the dependence structure into the joint probability distribution.
Such a structure is also independent of every covariate distribution, so it is usually represented as a function of a different set of variables $\{u_i\}$, that are referred to as \textit{transformed covariates}. Formally expressed, we have:
\[f(x_1,x_2,\ldots,x_d) = c(u_1,u_2,\ldots,u_d)\prod\limits_{i=1}^{d}f_i(x_i), \]
where $u_i = F_i(x_i).$
 
There are many families of parametric bivariate copula functions but only a few parametric $d$-variate ones.
Among the latter, Gaussian and T copulas are quite flexible because depend on the correlation matrix of the transformed covariates.

A better approach is to use bivariate copulas as building blocks for multivariate copulas.
This solution was first presented in \cite{Joe_1996}. 
Bedford and Cooke (\citeyear{Bedford_Cooke_2001}) later developed a more general factorization of multivariate densities and introduced regular vines.

\begin{definition}{(R-Vine on $d$ variables)}
\label{def:Rvine}
 A regular vine on $d$ variables consists of $d-1$ connected trees $T_1,...,T_{d-1}$, that satisfy the following:
 \begin{enumerate}
 	 \item $T_1$ consists of the node set $N_1=\{1,...,d\}$, where each variable is represented by only one of the nodes;
 	 and the edge set $E_1$, with each edge representing a copula that links two variables.
 	 \item For $i=2,\ldots,d-1$, the tree $T_i$ consists of the node set $N_i=E_{i-1}$ and the edge set $E_i$
 	 \item Each Tree $T_k$ has exactly $d-k$ edges, for $k=1,\ldots,d-1$. Two nodes in Tree $T_1$ can always form an edge in $T_1$. Two nodes in Tree $T_i$, with $i\ge2$, can form an edge in $T_i$ only if their corresponding edges in Tree $T_{i-1}$ share a common node.
 \end{enumerate}		  
\end{definition}

The regular vine copula $(\mathcal{V},\mathcal{B}_\mathcal{V},\theta_\mathcal{V})$ has density function defined as:

\[
c(u;\mathcal{V},\mathcal{B}_\mathcal{V},\theta_\mathcal{V})=\!\!\prod_{T_k\in\mathcal{V}}\prod_{e\in E_k}\!c_{\mathcal{B}_e}(u_{i(e)|D(e)},u_{j(e)|D(e)};\theta_e)
\]
where $\mathcal{B}_\mathcal{V}$ is the set of bivariate copula families selected for the Vine $\mathcal{V}$ and $\theta_\mathcal{V}$ the corresponding parameters.

For the sake of clarity, let us consider 5 variables $x_1,\ldots,x_5$, and that they have been already transformed via their corresponding marginals into $u_1,\ldots,u_5$, so $u_k = f_k(x_k)$.
According to definition \ref{def:Rvine}, a Tree $T_1$ will have 5 nodes and 4 edges. Therefore, its layout will necessarily be one out of those displayed in Figure \ref{fig:vine_layouts}(a--c). 
Each one of these layouts leads to many possible and different trees $T_1$, depending on how variables are arranged in the layout. 
For this example, let us assume that such an arrangement happens to be the one shown in the tree $T_1$ of Figure \ref{fig:vine_example}, in which it has been explicitly shown that the dependence (edge) between $u_3$ and $u_5$ is modeled with a Clayton copula, with parameter $\theta=1.7$. 
The layout of the next tree, $T_2$ may, or may not be one out of Figure \ref{fig:vine_layouts}(d--e). It depends on whether it satisfies the third requirement. 
In this example, the $T_1$ layout in Figure \ref{fig:vine_layouts}(a) imposes the layout in $T_2$ to be exclusively the one in Figure \ref{fig:vine_layouts}(d), resulting in the so called D-Vine. On the other hand, the $T_1$ layout in Figure \ref{fig:vine_layouts}(b) allows both Figures \ref{fig:vine_layouts}(d) and \ref{fig:vine_layouts}(e) to be layouts for $T_2$. 
When arranging the variables in the layout, it is a good practice to write the actual variables, and not the edges of the precedent tree. 
Then, all variables shared in both nodes of an edge are turned into conditioning variables, and the remaining are conditioned. 
Eventually, the factorization of the copula density is the product of all the nodes from $T_2$ on. 
Thus, copula density of the vine shown in Figure \ref{fig:vine_example} is
$$
\begin{array}{l}
c(u_1,\ldots,u_5)=  \\ \hspace*{1em}
c_{14}c_{24}c_{34}c_{35}
c_{13|4}c_{45|3}c_{23|4}
c_{15|34}c_{25|34}
c_{12|345}
\end{array}
$$
where the $c_{ij|k..}$ denotes the bi-variate copula density $c\big(F_i(x_i|x_k,..), (F_j(x_j|x_k,..)\big) $.

\section{Methodology}\label{method}
Model selection for vine is essentially a combinatorial problem over a huge search space, which is hard to solve with heuristic approaches. DiBmann proposed a tree-by-tree approach \cite{DiBmann_2013} , which selects maximum spanning trees at each level according to pairwise dependence. The problem with this locally greedy approach is that there is no guarantee that it will select a structure that is close to the optimal structure. 

In this section, we describe in details our learning approach for regular vine copula construction. In order to feed a vine configuration using a neural network based meta learning approach, we need to represent it in a compact way. Complexity arises since the construction of each level of tree depends on previous level's tree. The generated vine also needs to satisfy the following desirable properties:
\begin{itemize}
    \item \textbf{Tree Property} Each level in the vine should be a tree with no cycles.
    \item \textbf{Dependence Structure} The layout of each level tree depends on its previous level's tree.
    \item \textbf{Sparsity} In general, a sparse model where most of edges are independent copulas are preferred.
\end{itemize}
Next we compare two different representations for a regular vine model: Vector representation and RL representation.

\subsection{Vector Representation}
\begin{figure}[t]
    \centering
    \includegraphics[scale=0.5]{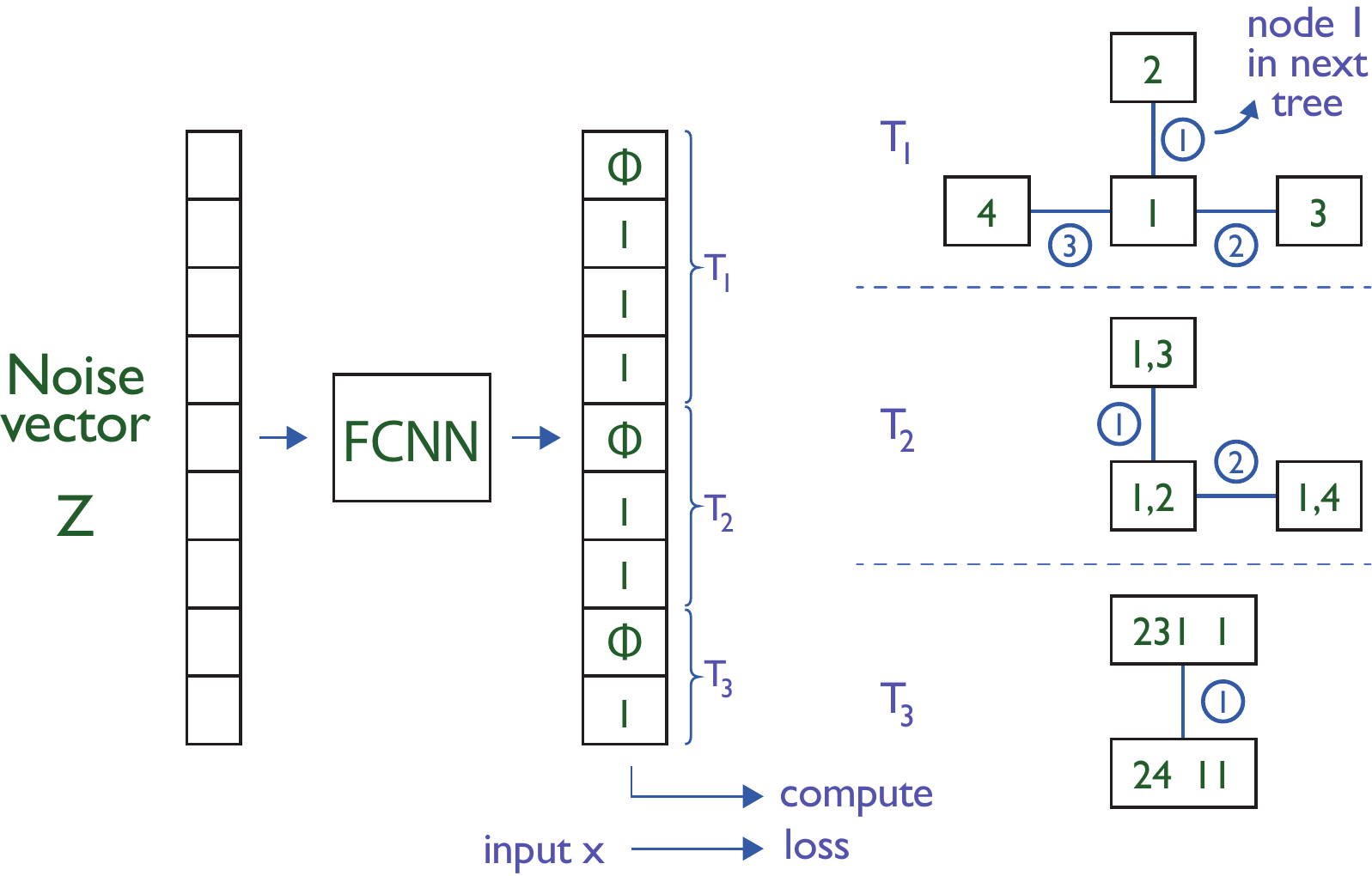}
    \caption{A example of the vector representation of vine. The model takes in a random initial configuration vector and search for the best vector that maximizes the objective function using a fully connected neural network (FCNN). The first element in the output vector (``$\Phi$'') means that first node in $T_1$ is not connected, the second element (``$1$'') means that second node in $T_1$ is connected to 1, etc. The layout on the right shows how to take the output vector and assembles it into a regular vine.}
    \label{fig:vector_representation}
\end{figure}

An intuitive way to embed a vine is to flatten all edges in the vine into a vector. Since the number of edges in the vine is fixed, the network can output a vector of edge indices. 
The representation proposed is depicted in Figure \ref{fig:vector_representation}.

Let the set of edges in each tree be $T_k$, the likelihood of the generated configuration can be computed as:
\[
\mathcal{L} = \sum_{T_k\in\mathcal{V}}\sum_{e\in E_k}\log c_{\mathcal{B}_e}(u_{i(e)|D(e)},u_{j(e)|D(e)};\theta_e))
\]
If the generated configuration contains a cycle at level $k$, a penalty will be subtracted from the objective function. The penalty will decrease with the level, indicating that since later trees depend on early trees, violation in tree property in early levels will incur a larger penalty. Let the number of cycles at level $k \in \{1\ldots K\}$ be $C_k$, thus the penalty due to violation of tree property can be computed as:
\[
J_1 = \frac{1}{k}\sum_{k=1}^K C_k
\]
In practice, most variables are not correlated and the vines tend to be sparse. To avoid over-fitting, we favor smaller edges in the vine graph by adding a penalty term $J_2$,  defined as:
\[
J_2 = \frac{1}{\sum_{k=1}^K|E_k|}
\]
At each training iteration, we are maximizing the following objective function:
\[
J_\phi = \mathcal{L}-\lambda J_1+\mu J_2
\]
where $\lambda$ and $\mu$ are hyper-parameters that can be tuned.

\subsection{Reinforcement Learning Representation}
One problem with the vector representation is that the vine configuration generated are not guaranteed to satisfy the tree property at each training step. 
On the other hand, the construction of the vine model can also be seen as a sequential decision making process:  At each step, the set of nodes $\mathcal{N}$ are partitioned into two sets, $\mathcal{N}^L$ and $\mathcal{N}^R$, where $\mathcal{N}^L$ denotes set of nodes that have been already added to the vine, and $\mathcal{N}^R$ denotes set of nodes that are not in the vine. When building tree $T_k$, in each step a node in $\mathcal{N}^R$ is selected and linked to a node in  $\mathcal{N}^L$, which is equivalent to adding a new edge to the tree $T_k$. Since pair of nodes already in the vine will never be selected at the same time, there won't be cycles forming, and therefore the tree property is maintained throughout the construction.

After we obtain the set of edges $E_k$ for tree $k$, we repeat the process for the next level of tree. The decision process can be defined as a fixed length sequence of actions of choosing $E_1,E_2,E_3,...$ where $|E_k|=d-k$. For an untruncated vine with $d$ variables, the total number of edges adds up to $\frac{d(d-1)}{2}$. 
Motivated by recent developments in RL, we reformulate the problem in its syntax.

\subsubsection{States}

Let $e_t$ be the $t$-th edge added to the vine model, which consists of a pair of indexes $(l_t,r_t)$. At step $t$, the states can be represented by the current set of edges in the vine $E_t$, and the partitioned vertices set $\mathcal{N}_t^L$ and $\mathcal{N}_t^R$. Formally $s_t=(E_t,\mathcal{N}_t^L$, $\mathcal{N}_t^R)$ is the state representation for the current vine at step $t$. 
\subsubsection{Actions}
The set of possible actions consists of all pairs of nodes that can be added to the current tree. 
Formally $A_s = \{(l_t,r_t):i_L \in \mathcal{N}^L, i_R \in \mathcal{N}^R\}$.
\subsubsection{Rewards}
The log likelihood of fitting a data point $\boldsymbol{x}$ to the vine $s_t$ at step $t$ can be decomposed into a set of trees and a set of independent nodes in $\mathcal{N}_t^R$ that have not been added to the tree:
\begin{align*}
\mathcal{L}(x,s_t)=& \sum_{e \in E_t,e=(l,r)}\hspace*{-1.2em}\log(c_{\mathcal{B}_e}(u_{l|D(e)},u_{r|D(e)};\theta_e)\\
&+\sum_{i \in \mathcal{N}^R_t}\log(u_{i})
\end{align*}
where $\mathcal{L}(x,s_0)=\sum_{v \in \mathcal{N}}log(u_v)$. 

Then the incremental reward at state $S_t$ can be defined as:
\begin{align*}
R_{likelihood}(S_t) &= \mathcal{L}(x,s_t)-\mathcal{L}(x,s_{t-1}) \\
     &= c_{B_{e_t}}(u_{l_t|D(e_t)},u_{r_t|D(e_t)};\theta_{e_t})\\
     & \hspace{1em} -\log(u_{r_t})
\end{align*}
where $e_t$ is the newly added edge to the vine. 

As before, we add a penalty term to avoid over-fitting.
Hence, the total reward is defined as:
\[
R = R_{likelihood} + \lambda R_{penalty}
\]

\subsubsection{Policy Network}
Let $\mathbb{P}_r$ be the true underlying distribution, and $\mathbb{P}_c$ be the distribution of the network.
Our goal is to learn a sequence of edges (pair of indices) $E$ of length $N-1$ such that the following objective function is maximized:
\[
J_\phi = \mathbb{E}_{\tau\sim \mathbb{P}_c}\mathbb{E}_{x\sim P_r}\left[\mathcal{L}(x,s_0)+\sum_{t=1}^{N-1}R(s_t)\right]
\]
Let $\pi_{\phi}$ be the stochastic policy defined implicitly by network $C_{\phi}$ and parameterized by $\phi$. 
The policy gradient of the above objective function $J_\phi$ can be computed as:
\[
\nabla_{\phi}J(\phi) = \mathbb{E}_{\tau\sim \mathbb{P}_c}\mathbb{E}_{x\sim P_r}\left[\sum_{t=1}^{N-1}R(s_t)\nabla_{\phi}log\pi_{\phi}(\tau_t|\hat{s}_{t-1})\right]
\]
At each step, the policy network outputs a distribution over all possible actions, from which an action (edge) is sampled.
Following standard practice, the expectation is approximated by sampling $n$ data points and $m$ action sequences per data point. Additionally, we use discounted reward and subtract a baseline term to reduce variance of gradients \cite{greensmith}.

\begin{algorithm}
\caption{Vine Learning}\label{learning}
\begin{algorithmic}
\Procedure{Training}{}
\For{\text{number of training iterations}}
\State Sample m examples $\{x^{(1)},...,x^{(m)}\} \in \mathbf{R}^d${\scriptsize$_\rightarrow$}
\State \hfill{\scriptsize$_\rightarrow$}from real distribution
\State Transform examples into $\{F_1(x_1^{(1)}),...,F_d(x_d^{(1)})\}$
\State $E = \emptyset, \mathcal{N}^L = \emptyset, \mathcal{N}^R = \mathcal{N}$
\While{$|E| <$\text{total num of edges}}
\State sample action $i,j \sim \pi(a|s_t)$
\State Find $\mathcal{B}^*,\theta^*$ for edge $\{i,j\}$
\State $\mathcal{N}^L = \mathcal{N}^L \bigcup \{i\}$, $\mathcal{N}^R = \mathcal{N}^R \setminus \{j\}$ 
\State $E_t = E_{t-1}\bigcup \{i,j\} $
\State Calculate step reward $R_t$
\EndWhile
\State update the model by descending its {\scriptsize$_\rightarrow$}
\State \hfill {\scriptsize$_\rightarrow$} stochastic gradient $\nabla_{\phi}J(\phi)$
\EndFor
\EndProcedure{}
\end{algorithmic}
\end{algorithm}

When constructing the Vine, a decision made in the first tree can affect to the nodes in deeper trees. Therefore, we cannot assure the markovian property, i.e. that the next state depends only of the current state and current decisions. A natural choice will be to adopt LSTM as a solution to this non-Markovian reinforcement learning tasks with long-term dependencies. Once the configuration is determined, we can find copula family and parameter for each edge following the method described in the following section.

\begin{figure}[t]
    \centering
    \includegraphics[scale=0.52]{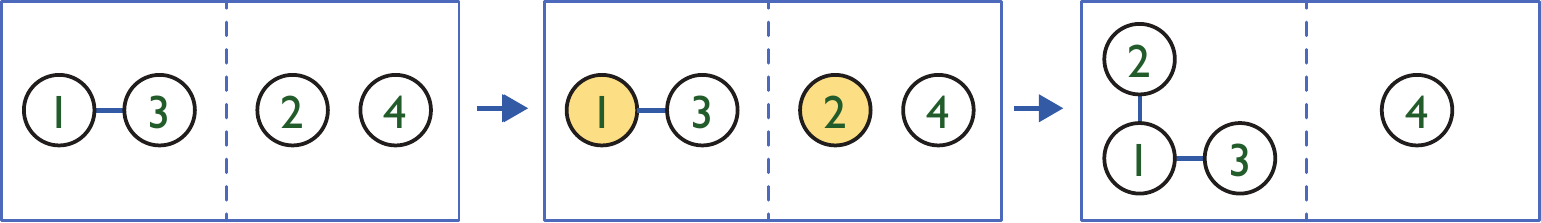}
    \caption{A example of a training step for a 4-dim vine with RL formulation. The set of nodes is partitioned into $\mathcal{V}^L = \{1,3\}$ and $\mathcal{V}^R = \{2,4\}$. Node 1 in $\mathcal{V}^L$ and node 2 in $\mathcal{V}^R$ is sampled by the policy network and edge $\{1,2\}$ is added to the vine. }
    \label{fig:rl_rep}
\end{figure}

\subsection{Pair Copula Selection}
For each edge, we estimate the log-likelihood of the empirical density, and both the fit to the left tail and to the right tail. Then we combine these three measurements in a hard-voting fashion to select a bivariate copula family according to the three-way check methods \cite{Alfredo_Kalyan_2015}. The parameter is estimated according to its max likelihood fit after fitting the bivariate copula family. This provides us an approximated reward for adding an edge that guides us through the search space.

\subsection{Sampling From A Vine}
\begin{figure}[h]
    \centering
    \includegraphics[width=.9\linewidth]{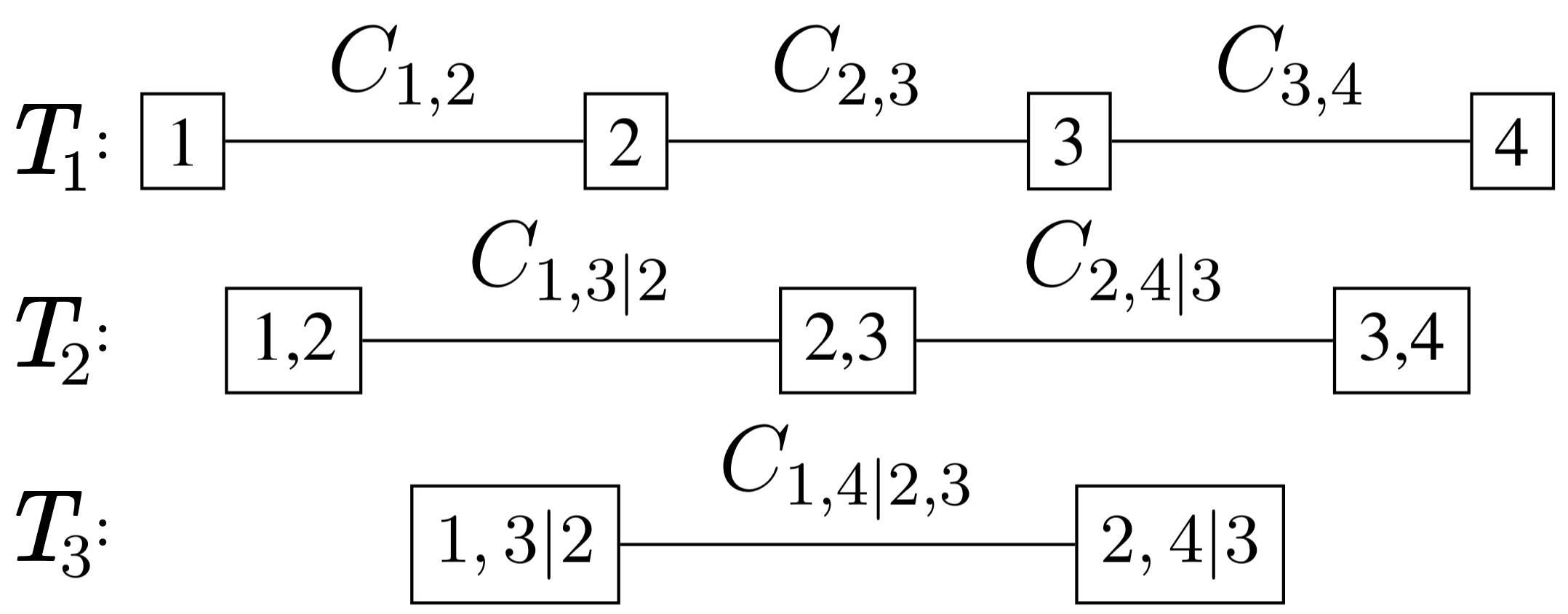}
    \caption{D-Vine of 4 variables.}
    \label{fig:Dvine}
\end{figure}

After learning a vine model from the data, we can sample synthetic data from it.
Kurowicka and Cooke first proposed an algorithm to sample an arbitrary element from a regular vine, which they call the Edging Up Sampling Algorithm \cite{Kurowicka_Cooke_2007}. 
The sampling procedure requires a complete vine model of $n$ nodes $X_1,X_2,...,X_n$, and their corresponding marginal distributions $F_1,F_2,...,F_n$.
Considering the case where we have a D-Vine model with 4 nodes, shown in Figure \ref{fig:Dvine}. 
We start by sampling univariate distributions $u_1,u_2,u_3,u_4$ from Uniform Distribution over [0,1]. We randomly pick a node to start with, say $X_2$.

Then the first variable $x_2 \sim X_2$ can be sampled as:
\begin{align}
x_2 = F_2^{-1}(u_2)
\end{align}
After we have $x_2$, we randomly pick a node connected to $X_2$. Suppose we pick $X_3$, recall that the conditional density $f_{3|2}$ can be written as:
\begin{align}
f_{3|2}(x_3|x_2)&=f_3(x_3)c_{2,3}(F_2(x_2),F_3(x_3)) \\
&=f_3(x_3)c_{2,3}(u_2,u_3) \\
&=f_3(x_3)c_{3|2}(u_3)
\end{align}
Thus, $x_3 \sim X_3|X_2$ can be sampled by:
\begin{align}
x_3 = F_3^{-1}(C_{3|2}^{-1}(u_3))
\end{align}
where $C_{3|2}$ can be obtained from $C_{2,3}$ in $T_1$ by plugging in sampled values of $u_2$. 

Similarly, we pick a node that shares an edge with $X_3$, say $X_4$. Then $x_4 \sim X_4|X_2,X_3$ can be sampled as:
\begin{align}
x_4= F_4^{-1}\circ C_{4|3}^{-1}\circ C_{4|23}^{-1}(u_4)
\end{align}
Finally $x1 \sim X_1|X_2,X_3,X_4$ can be sampled as:
\begin{align}
x_1= F_1^{-1}\circ C_{1|4}^{-1}\circ C_{1|34}^{-1}\circ C_{1|234}^{-1}(u_1)
\end{align}

For a more general vine graph, we can use a modified Breadth First Search to traverse the tree and keep track of nodes that have already been sampled. The general procedure is described in Algorithm \ref{sampling}.
\begin{algorithm}
\caption{Sampling from a regular vine}\label{sampling}
\begin{algorithmic}
\Procedure{Sampling}{}
\State $\textit{explore} \gets \text{Queue()}$
\State $\textit{visited} \gets \text{list()}$
\State $start \gets \text{randomly chosen start node}$
\State \textit{explore}.\text{enqueue(start)}
\While {\text{explore not empty}}
\State $cur \gets \textit{explore.}\text{dequeue()}$
\State $i \gets \text{len(visited)}$
\State $x_{cur}=F_{cur}^{-1}\circ C_{cur|visited[i-1]}^{-1}...$
\State \hspace{1.2cm}$\circ ~ C_{cur|visited[1,..,i-1]}^{-1}(u_{cur})$
 \For{\text{s $\in$ neighbor(cur)}}
        \If{s $\in$ visited}
                \State Continue
        \Else
        		\State \textit{explore}.\text{enqueue(s)}
        \EndIf
\EndFor
\State \textit{visited.}\text{left\_append(\textit{cur})} 
\EndWhile
\EndProcedure
\end{algorithmic}
\end{algorithm}

\section{Experiments}\label{experiments}
\subsection{Baselines and Experiment Setup}
We compare our models with the tree-by-tree greedy approach \cite{DiBmann_2013}, which selects maximum spanning trees at each level according to pairwise dependence, as well as the bayesian approach \cite{Gruber_Czado_2015}. To test our approach, we used constructed examples. In this we construct vines manually, sample data from them and then use our learning approach to recreate the vine. We use two of the constructed examples defined in 
\cite{Min_Czado_2010} and compare the results with both the greedy approach and the Bayesian approach.
Second, we used three real data sets and compared the two approaches. The neural network used for creating vines for the vector representation is set-up as fully connected feed forward neural networks with two hidden layers. Each layer uses ReLU as activation function and the output layer is normalized by a softmax layer. The network for the reinforcement learning representation is set up as LSTM. The algorithm is trained over 50 epochs in the experiments. 

\subsection{Constructed Examples}
To illustrate a scenario where the greedy approach might fail, we independently generate data sets of size 500 from a pre-specified regular vine. The fitting abilities of each model are measured by the ratio of log-likelihood of the estimated model and log-likelihood of the true underlying model. 
\begin{itemize}
    \item \textbf{Dense Vine} Dense Vine is a  6-dim vine where all bivariate copulas are non-independent. The bivariate families and thetas are listed in Table \ref{tab:dense_vine}.
    \item \textbf{Sparse Vine} A 6-dim vine where all trees consisted of independent bivariate copulas except the first tree. In other words, the vine is truncated after the first level.
\end{itemize}

\begin{table}[h!]
    \centering
    \label{table1}
    \caption{Comparison of relative log-likelihood on constructed examples. All results reported here are based on 100 independent trails.}
    \begin{tabular}{llll}
    \hline
                            & Dense & Sparse & Dense \\ & & &T1 correct\\ \hline
    Di{\ss}mann relative loglik(\%) & 76.6 & \textbf{101.3}& No\\ \hline
    Gruber relative loglik(\%) & 81.0 & 100.6& No \\ \hline
    Vector relative loglik(\%) &80.3 &99.8 & No \\ \hline
    RL relative loglik(\%) & \textbf{84.2} & 100.2& \textbf{Yes}\\   \hline
    \end{tabular}
\end{table}
In the dense vine example, the RL vine is the only model able to recover the correct first tree and also achieves the highest relative log-likelihood. In the sparse vine example, all four models achieve similar results, among which Di{\ss}mann obtains the highest likelihood. As argued in \cite{Min_Czado_2010}, the higher likelihood in Di{\ss}mann is achieved at the expense of over-fitting. The two examples demonstrate the fitness of the model is improved by our model under different scenarios. Moreover, for a 6-dimensional data set of size 500, the RL algorithm finishes in approximately 15 minutes with a single GPU.

\subsection{Real Data}
In this section, we apply vine model to real data sets for the task of synthetic data generation. The three data sets that are picked are a binary classification problem, a multi-class classification problem and a regression problem. The batch size used is 64 for breast cancer dataset and 128 for the other two datasets. The three data sets used in experiments are:
\begin{itemize}
     \item \textbf{Wisconsin Breast Cancer} Describes 30 variables computed from a digitized image of a fine needle aspirate (FNA) of a breast mass and a binary variable indicating if the mass is benign or malignant. This dataset includes 569 instances.
    \item \textbf{Wine Quality} This dataset includes 11 physiochemical variables and a quality score between 0 and 10 for red (1599 instances) and white (4898 instances) variants of the Portuguese "Vinho Verde" wine.
    \item \textbf{Crime} The communities and crime dataset includes 100 variables related to crimes ranging from socio-economic data to law enforcement data and an attribute to be predicted (Per Capita Violent Crime) . This dataset has 1994 instances.
\end{itemize}

To evaluate the quality of the synthetic data, we first evaluate its log likelihood. As shown in \ref{table2}, synthetic data generated from RL Vine achieves the highest log likelihood per instance in all three datasets. 

\begin{table}[h!]
\centering
\caption{log-likelihood per instance truncated after third tree}
\label{table2}
\begin{tabular}{llll}
\hline
            & Breast Cancer & Wine & Crime  \\ \hline
Di{\ss}mann & 0.79  & 0.033  &  0.26       \\ \hline
Vector Rep & 0.84  & 0.037  &  0.27         \\ \hline
RL Vine   &   \textbf{0.91}  & \textbf{0.045}  &  \textbf{0.31}             \\ \hline

\end{tabular}
\end{table}

Besides log-likelihood, the quality of the generated synthetic data is also evaluated from a more practical perspective. High quality synthetic datasets enable people to draw conclusions and make inferences as if they are working with a real data set. We first use both Di{\ss}mann's algorithm and our proposed algorithms to learn copula vine models from a real data set. Later, we generate synthetic data from each model and train models for target variables on the synthetic training set as well as real training set, and use the real testing set to compute the corresponding evaluation metric (F1 score for classification and MSE for regression).  For ease of computation, the learned vines are truncated after the third level, which means all pair copulas are assumed to be independent beyond the third level. All results reported are based on 10-fold cross-validation over different splits of training and testing set. 

As shown in table \ref{table2} and table \ref{table3}, synthetic data from RL Vine achieves highest F1 score and the results are comparable to real data. For regression data, table \ref{table4} demonstrates that synthetic data obtains lowest Mean Squared Error (MSE) among the models. The results shown demonstrate that our proposed model improve the overall model selection and is able to generate reliable synthetic data.

\begin{table}[h!] \footnotesize
\centering
\caption{F1 score of different end classifiers on breastcancer dataset}
\label{table3}
\begin{tabular}{lllll}
\hline
            & Decision Tree & SVM & 5 layer MLP \\ \hline
Real Data   &   $0.92 \pm 0.01$ & $0.90 \pm 0.02$   &   $0.93\pm 0.02$               \\ \hline
Di{\ss}mann &  $0.77 \pm 0.12$ & $0.55\pm 0.05$    &    $0.76\pm 0.06 $         \\ \hline
Vector Rep &  $0.76 \pm 0.09$          &  $0.62 \pm 0.07$  &    $ 0.71 \pm 0.05$             \\ \hline
RL Vine   &     $\textbf{0.81} \pm 0.06$          &  $\textbf{0.72} \pm 0.03$  &    $ \textbf{0.79} \pm 0.03$              \\ \hline

\end{tabular}
\end{table}

\begin{table}[h!]  \footnotesize
\centering
\caption{F1 score of different end classifiers on wine quality datset (averaged over 11 classes)}
\label{table4}
\begin{tabular}{lllll}
\hline
            & Decision Tree & SVM  & 5 layer MLP \\ \hline
Real Data   &   $0.36 \pm 0.016 $ & $0.11 \pm 0.015 $    &   $0.16 \pm 0.016$              \\ \hline
Di{\ss}mann &  $0.13 \pm 0.008$ & $0.09 \pm 0.006$    &    $0.09 \pm 0.006 $        \\ \hline
Vector Rep & $0.15 \pm 0.011$ & $0.07 \pm 0.004$    &    $0.08 \pm 0.005 $        \\ \hline
RL Vine   & $\textbf{0.21} \pm 0.006$ & $\textbf{0.09} \pm 0.006$    &    $\textbf{0.11} \pm 0.008 $          \\ \hline

\end{tabular}
\end{table}

\begin{table}[h!]   \footnotesize
\centering
\caption{MSE error of different end classifiers on crime dataset}
\label{table5}
\begin{tabular}{lllll}
\hline
            & Decision Tree & SVM &  5 layer MLP \\ \hline
Real Data   & $0.045 \pm 0.004 $   &  $0.021 \pm 0.001 $   &   $0.023\pm 0.001$                    \\ \hline
Di{\ss}mann & $0.137 \pm 0.016$  & $0.081\pm 0.009$   &  $0.124\pm0.053$                \\ \hline
Vector Rep &  $0.112 \pm 0.014$  & $0.075\pm 0.006$   &  $0.116\pm0.021$           \\ \hline
RL Vine   &  $\textbf{0.096} \pm 0.011$  & $\textbf{0.072}\pm 0.007$   &  $\textbf{0.109}\pm0.032$  \\ \hline

\end{tabular}
\end{table}

\section{Conclusion}
In this paper we presented a meta learning approach to create vine models for modeling high-dimensional data. Vine models allow for the creation of flexible structures using bivariate building blocks. However, to learn the best possible model, one has to identify the best possible structure, which necessitates identifying the connections between the variables and selecting between the multiple bivariate copulas for each pair in the structure. We formulated the problem as a sequential decision making problem similar to reinforcement learning, used long-short-term memory networks to simultaneously learn the structure and select the bivariate building blocks. We compared our results to the state of the art approaches and found that we achieve significantly better performance across multiple data sets. We also show that our approach can generate higher quality synthetic data that could be directly used to learn a machine learning model replacing the real data.


\section*{Acknowledgment}
Dr. Cuesta-Infante is funded by the Spanish Government Research Project TIN-2015-69542-C2-1-R (MINECO/FEDER)  and the Banco de Santander grant for the Computer Vision and Image Processing Excellence Research Group (CVIP). Dr. Kalyan Veeramachaneni and Yi Sun acknowledge the generous funding provided by Accenture and National Science Foundation under the grant ``\textit{CIF21 DIBBs: Building a Scalable Infrastructure for Data-Driven Discovery and Innovation in Education}'', Award \# 1443068.

\bibliographystyle{aaai}
\bibliography{bibfile,copulasALF}

\end{document}